%% file: main.tex
\documentclass[conference]{IEEEtran}
\IEEEoverridecommandlockouts

\usepackage{cite}
\usepackage{amsmath,amssymb,amsfonts}
\usepackage{algorithmic}
\usepackage{float}
\usepackage{subcaption}
\usepackage{textcomp}
\usepackage{caption}
\usepackage{xcolor}
\usepackage{graphicx}
\usepackage{hyperref}

\newcommand{\NAME}{\textsc{SimPRIVE}}

\def\BibTeX{{\rm B\kern-.05em{\sc i\kern-.025em b}\kern-.08em
    T\kern-.1667em\lower.7ex\hbox{E}\kern-.125emX}}
\begin{document}

\title{SimPRIVE: a Simulation framework for Physical Robot Interaction with Virtual Environments
}

\author{\IEEEauthorblockN{Federico Nesti, Gianluca D'Amico, Mauro Marinoni, Giorgio Buttazzo}
\IEEEauthorblockA{\textit{Department of Excellence in Robotics \& AI} \\
\textit{Scuola Superiore Sant'Anna}, Pisa, Italy \\
$<$name$>$.$<$surname$>$@santannapisa.it}
}

%


\maketitle



\begin{abstract}
The use of machine learning in cyber-physical systems has attracted the interest of both industry and academia. However, no general solution has yet been found against the unpredictable behavior of neural networks and reinforcement learning agents.
Nevertheless, the improvements of photo-realistic simulators have paved the way towards extensive testing of complex algorithms in different virtual scenarios, which would be expensive and dangerous to implement in the real world. 

This paper presents \NAME, a simulation framework for physical robot interaction with virtual environments, which operates as a vehicle-in-the-loop platform, rendering a virtual world while operating the vehicle in the real world. 

Using \NAME, any physical mobile robot running on ROS 2 can easily be configured to move its digital twin in a virtual world built with the Unreal Engine 5 graphic engine, which can be populated with objects, people, or other vehicles with programmable behavior.

\NAME~has been designed to accommodate custom or pre-built virtual worlds while being light-weight to contain execution times and allow fast rendering. Its main advantage lies in the possibility of testing complex algorithms on the full software and hardware stack while minimizing the risks and costs of a test campaign.
The framework has been validated by testing a reinforcement learning agent trained for obstacle avoidance on an AgileX Scout Mini rover that navigates a virtual office environment where everyday objects and people are placed as obstacles. The physical rover moves with no collision in an indoor limited space, thanks to a LiDAR-based heuristic.
\end{abstract}

\vspace{2mm}
\begin{IEEEkeywords}
Simulation environment, ROS, Cyber-physical systems, Digital Twins, Reinforcement Learning
\end{IEEEkeywords}

\begin{figure}[!ht]
  \centering
  \includegraphics[trim=110 0 100 0, clip,width=0.49\columnwidth]{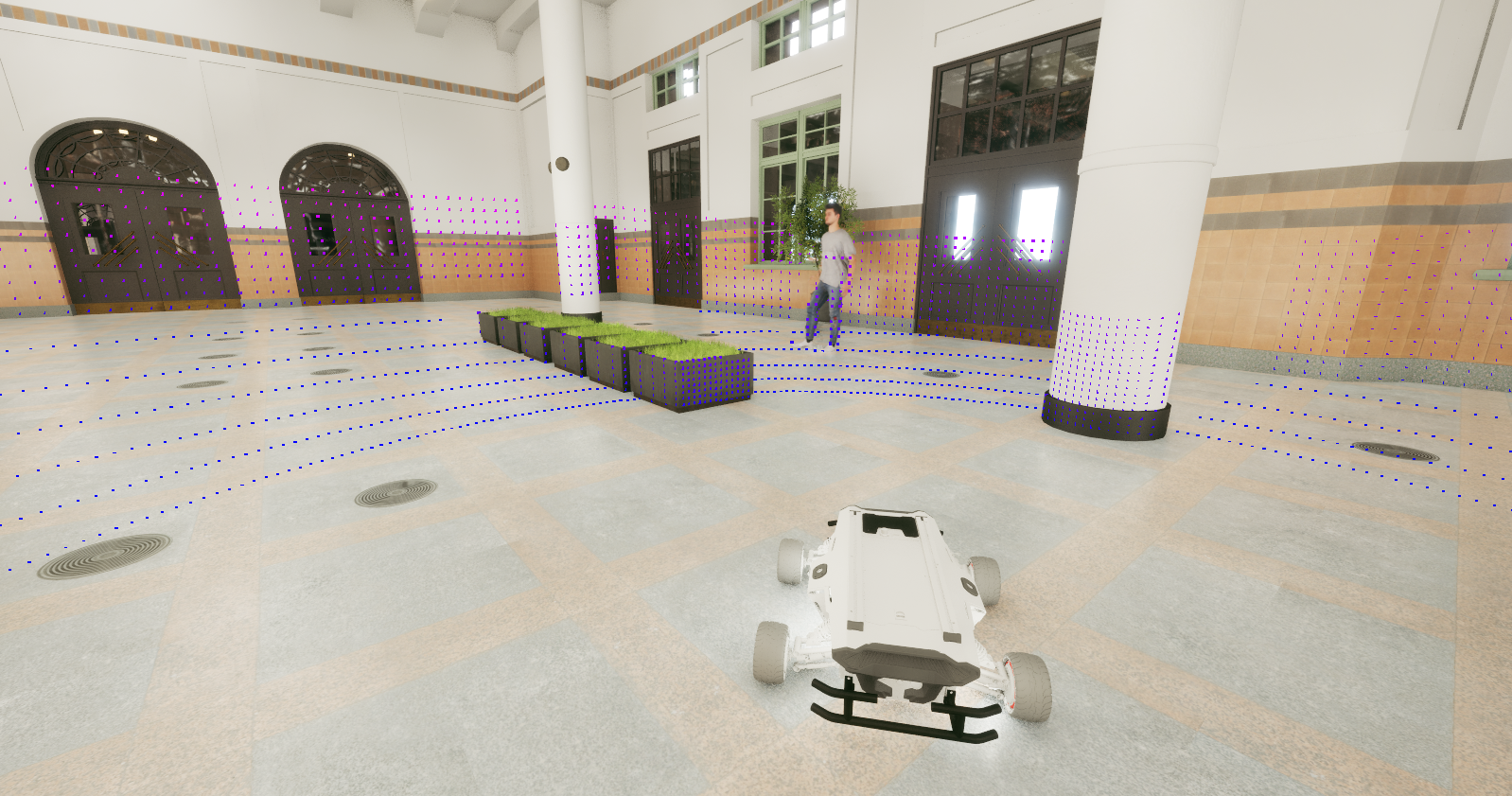}
  \includegraphics[width=0.49\columnwidth]{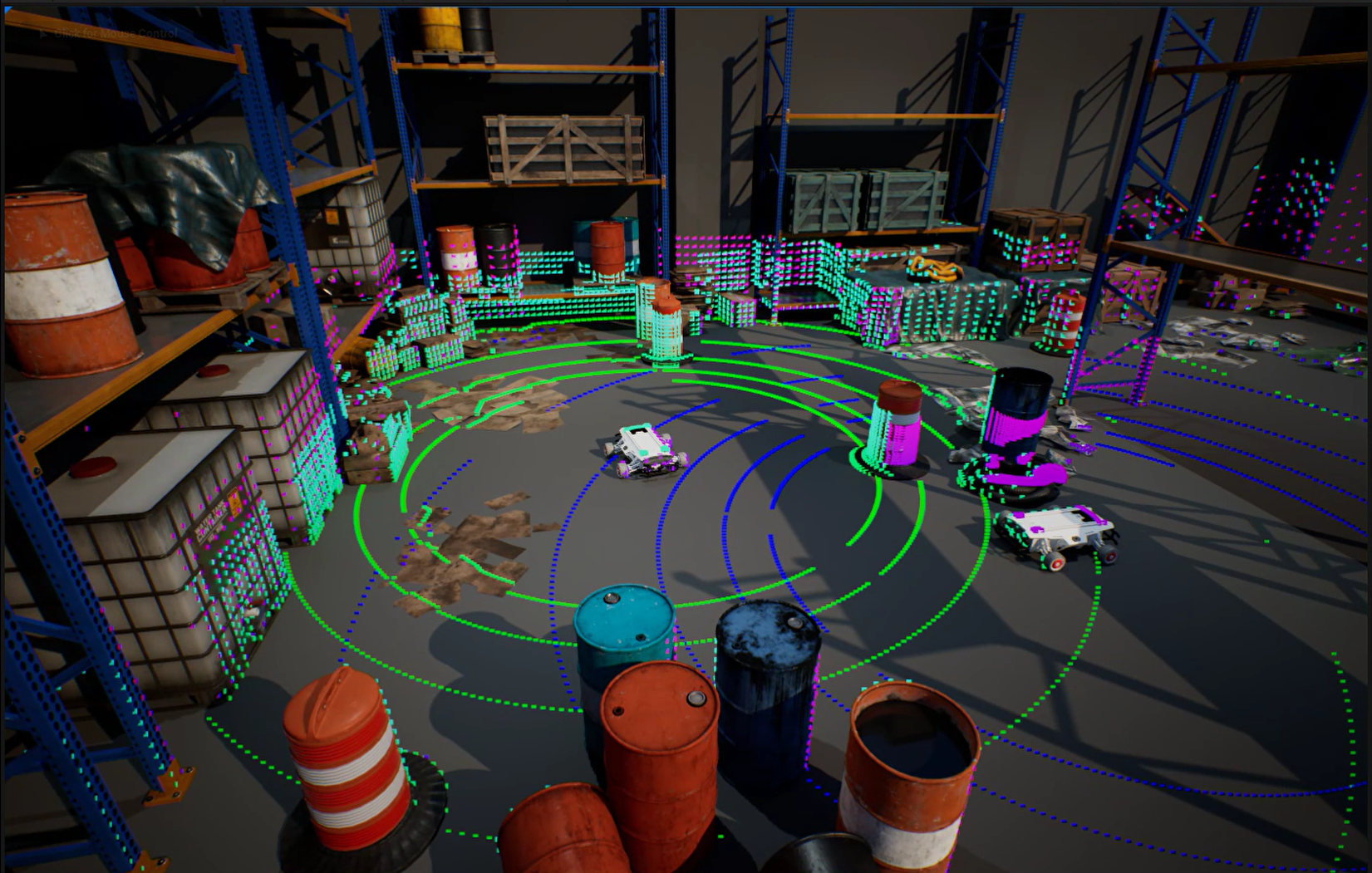}
  \includegraphics[width=0.49\columnwidth]{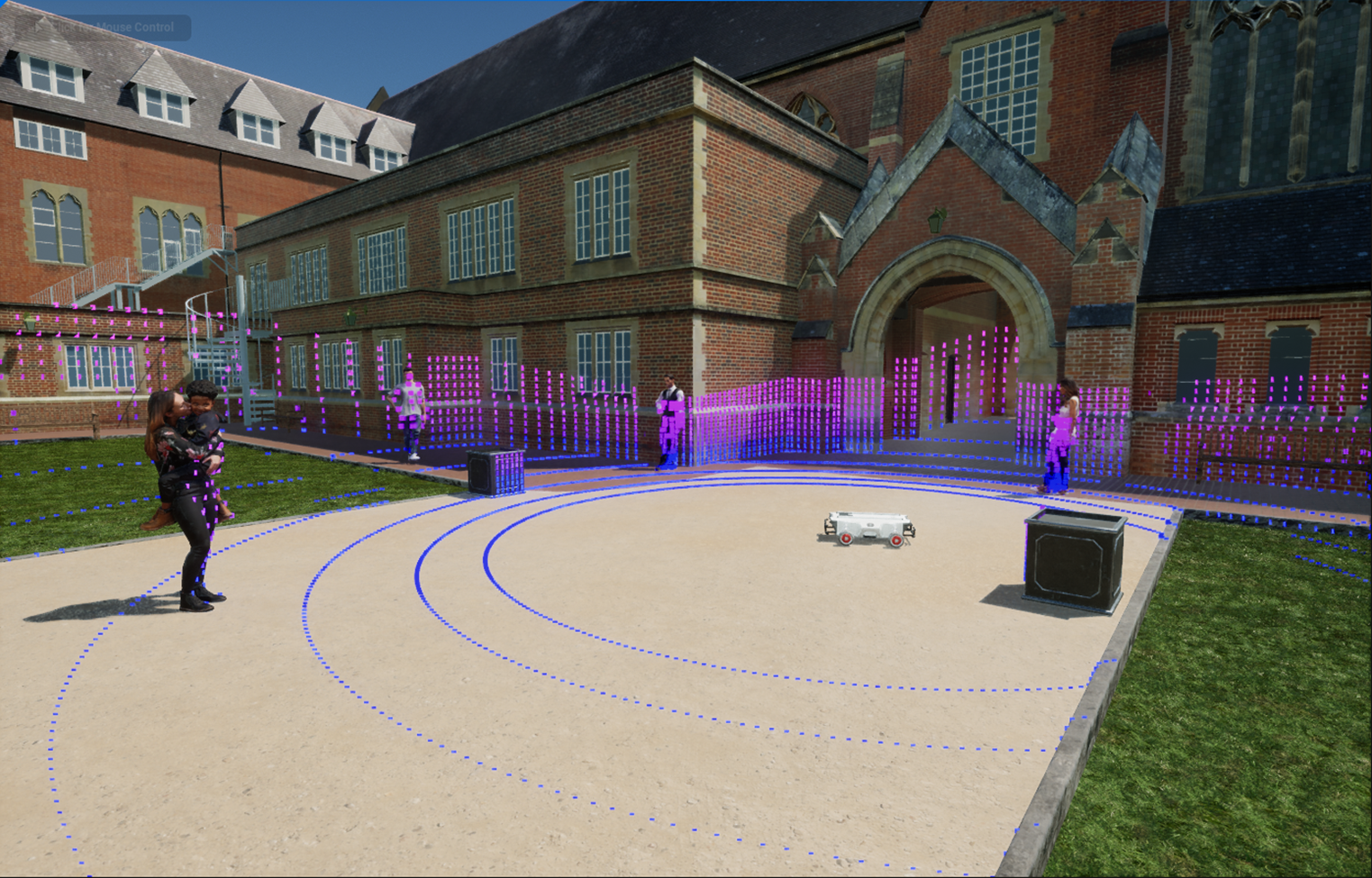}
  \includegraphics[width=0.49\columnwidth]{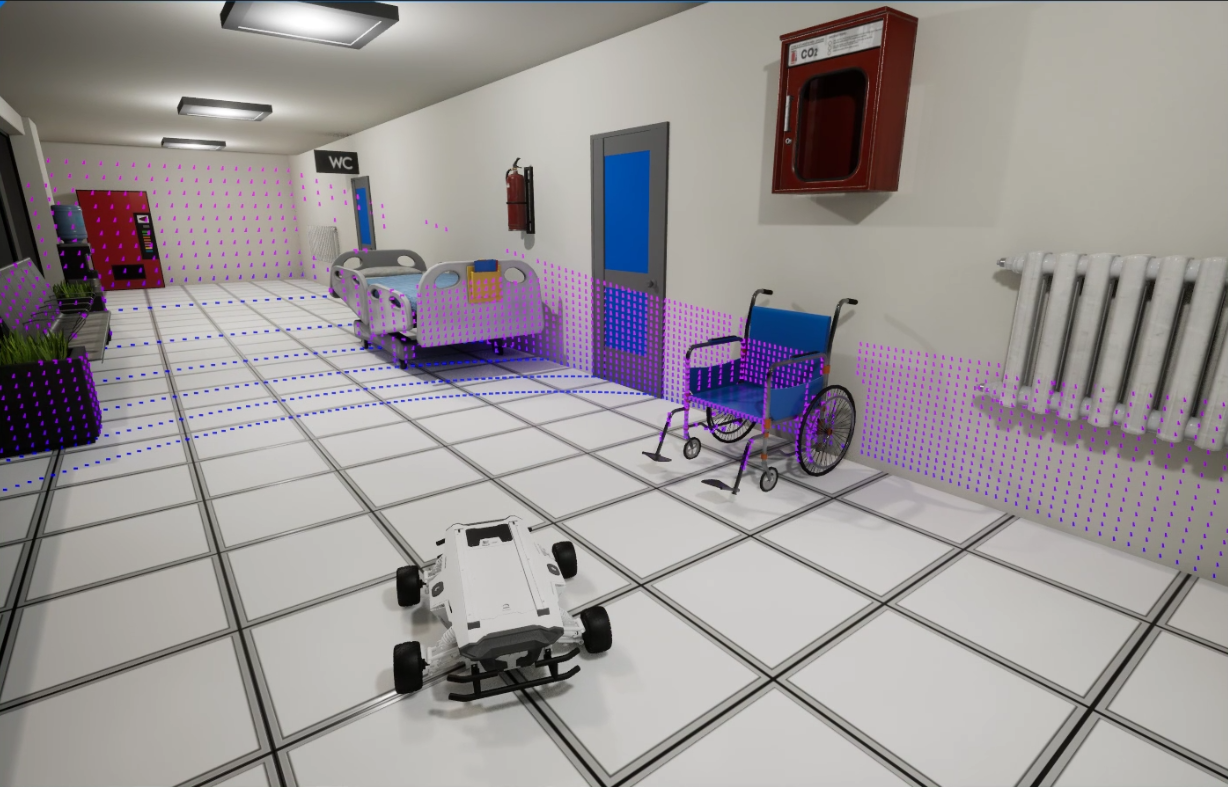}
  \includegraphics[width=0.6\columnwidth]{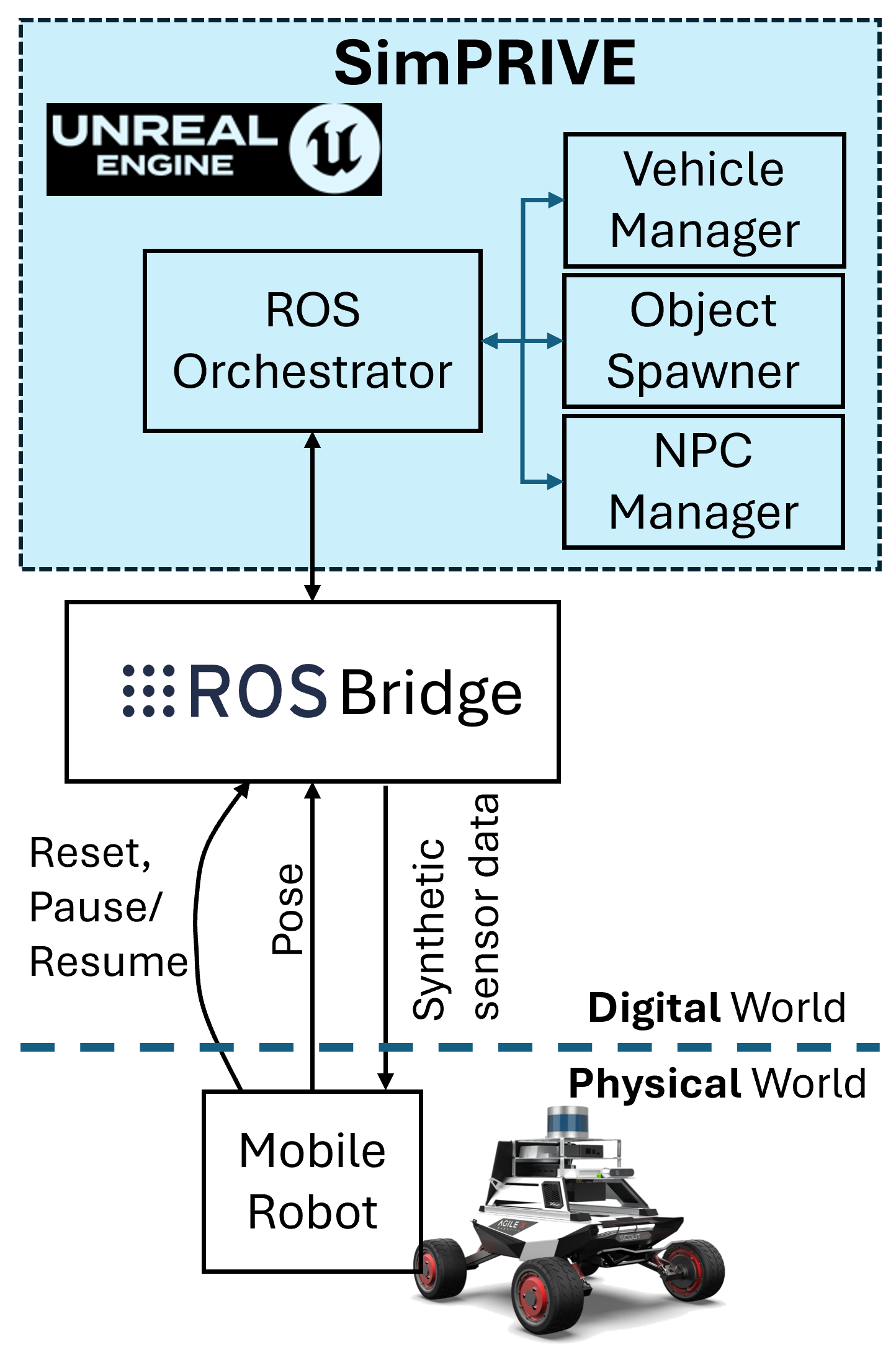}
  \caption{\small High-level overview of the architecture of the proposed \NAME~framework. The figure also shows renderings of the digital twin of the rover in different virtual environments, with synthetic LiDAR data superimposed. }
  \label{fig:fig1}
\end{figure}

\input{01-intro}

\input{02-related}
\input{03-method}
\input{04-experiments}

\input{05-conclusion}

\section*{Acknowledgment}
{\small 
This work was partially supported by project SERICS (PE00000014) under the MUR National Recovery and Resilience Plan funded by the European Union - NextGenerationEU.}

\bibliographystyle{IEEEtran}
\bibliography{bibliography}

\input{suppl}

\end{document}

%% file: 01-intro.tex
\section{Introduction}\label{s:intro}

Recently, artificial intelligence (AI) models have achieved impressive performance in many applications, including cyber-physical systems (CPS).
In particular, autonomous vehicles are becoming a reality, and deep neural networks (DNNs) are the de facto standard for perception tasks.
However, including AI in CPS presents several hurdles that must be overcome before deploying AI-based controllers in the wild~\cite{rech2024artificial},   \cite{perez2024artificial}.
%
One of the main issues around DNNs is their unpredictable behavior, which might occur in rare situations in which out-of-distribution or adversarial inputs are presented to the model.
To overcome such a problem, different approaches have been proposed in the literature, including dedicated architectures~\cite{8897630}, model verification/robustness certificates~\cite{wang2024survey}, \cite{meng2022adversarial}, and adversarial/out-of-distribution detection~\cite{aldahdooh2022adversarial}, \cite{yang2024generalized}, \cite{salehi2021unified}. 

At the same time, the last few years have seen an incredible improvement and democratization of photo-realistic simulators based on open-source graphics engines such as Unity~\cite{unity} and Unreal Engine~\cite{ue5}.
Exploiting the capabilities of a photo-realistic simulator offers the opportunity to safely and extensively test the system under development in many different situations\cite{10576049}, reducing risks, costs, and time for running expensive experiments required to optimize sensors configuration and algorithms.

In particular, the hardware-in-the-loop (HIL)~\cite{mihalivc2022hardware} approach is based on the execution of the real-time software stack directly on the target platform (typically, an embedded system), while the dynamics of the system and its sensors are simulated with a mathematical model (digital twin~\cite{tao2022digital}).

For a vehicle, this concept can be pushed further by considering not only the software stack, but the entire hardware, resulting in a vehicle-in-the-loop simulation~\cite{cheng2023survey}. This enables the vehicle to operate in the real world, while its digital twin moves accordingly in the virtual world, perceiving the virtual reality rendered by a graphics engine. This setup allows testing the algorithms when deployed on the actual target hardware, while safely reconstructing situations that would be dangerous, risky, or expensive to replicate in the real world (e.g., close interactions with people or other moving vehicles). 

Inspired by this concept, this paper presents \NAME, a \textbf{Sim}ulation framework for \textbf{P}hysical \textbf{R}obot \textbf{I}nteraction with \textbf{V}irtual \textbf{E}nvironments built using Unreal Engine 5 (UE5). 
\NAME\ can be configured to communicate with a physical robot through the Robot Operating System (ROS) 2, which is widely adopted in industry and robotics research.
Specifically, the framework is designed to provide a straightforward integration with any ROS2-enabled terrestrial vehicle, whose digital twin (i.e., its 3D mesh and simulated sensors) is moved in a virtual world according to its movements in the real world.

The framework is programmed to generate synthetic sensor readings, such as cameras and LiDARs, which are provided as inputs to the algorithms under test. 
While such algorithms are safely tested in the virtual world, where the robot's digital twin operates, the physical robot moves in the real world and must be equipped with an obstacle detection algorithm (processing physical sensors) to prevent collisions with physical objects.

Accessing the state of both the physical robot and its virtual counterpart is extremely useful non only to safely test the behavior of complex algorithms, but also to fine-tune AI-based algorithms directly in a virtual environment, while the vehicle operates in the real world according to its real dynamics.

The framework was tested on a use-case where an AgileX Scout Mini rover~\cite{scout_mini} performs camera- and LiDAR-based navigation and obstacle avoidance in a virtual environment that replicates a corridor in our laboratory. However, the framework is flexible enough to simulate any ROS2-enabled mobile robot and any custom or predefined virtual environment. 
To summarize, this paper presents the following contributions:
\begin{itemize}
    \item \NAME, a flexible simulation framework for Physical Robot Interaction with Virtual Environments, which generates sensor readings from a digital world as inputs to a physical robot in the real world.
    \item A case study for an AI-based controller deployed on a real rover, tested using \NAME. Our custom implementation showcases the potential of the framework. 
\end{itemize}

The paper is organized as follows: Section~\ref{s:related} presents the related literature, Section~\ref{s:method} provides implementation details of the framework, Section~\ref{s:exp} shows the experimental results, and Section~\ref{s:conclusions} states the conclusions, discusses the limitations, and illustrates future directions.

%% file: 02-related.tex
\section{Background and Related Works}\label{s:related}

\NAME~is inspired by the vehicle-in-the-loop simulation paradigm, which has its roots in the hardware-in-the-loop~\cite{mihalivc2022hardware} simulation (HIL). HIL relies on a mathematical model of the system under test and requires the software stack to be executed on the target embedded real-time board, which can, therefore, be tested in different conditions without ever leaving the test bench. 
After being introduced in NASA's Apollo missions~\cite{nasa_digital_twin}, the concept of digital twin has been refined as a complex, dynamic virtual entity that reflects the current state and condition of its real-world counterpart~\cite{9670454}. Digital twins have effectively been used to simulate, analyze, and optimize~\cite{botin2022digital} all kinds of physical processes. 

The vehicle-in-the-loop paradigm is strictly linked to the digital twin concept and is specifically used to test key sub-systems~\cite{9361649} or the behavior of the entire vehicle~\cite{cheng2023survey}. This technology is incredibly useful in the autonomous driving domain and has been implemented in the industry and in several scientific works. 
For instance, Shen et al. presented Sim-on-wheels~\cite{shen2023simonwheels}, a vehicle-in-the-loop simulation framework for autonomous driving that is able to add virtual entities to the real images to obtain mixed-reality renderings; however, the framework focuses on cameras and does not support LiDAR.
Wang et al.~\cite{10057099} investigated the vehicle- and pedestrian-in-the-loop co-simulation, using the Cave automatic virtual environment and the Carla simulator~\cite{dosovitskiy2017carla}. 
Xiong et al.~\cite{xiong2022design} proposed a vehicle-in-the-loop car following simulation framework built on Unity, the same graphics engine used by Wang et al.~\cite{wang2021digital} to simulate the hardware of connected vehicles.
In other works, the vehicle-in-the-loop framework is used to reduce the gap between simulation and reality while testing specific methodologies, such as reinforcement learning~\cite{VOOGD20231510} and nonlinear control~\cite{allamaa2022sim2real}.
Similarly to the setting proposed in this paper, Hiba et al.~\cite{10156429} explored the vehicle-in-the-loop paradigm for drones using ROS and Carla.

Most of the vehicle-in-the-loop simulators described above are explicitly designed for autonomous driving and require expensive hardware to work. Conversely, \NAME~is a general simulation framework that can work with any ROS2-enabled mobile robot, thus allowing the user to test a more extensive set of algorithms on a broader range of platforms in terms of sensors and computational capabilities.

Furthermore, while most of the previous works were primarily based on off-the-shelf simulators such as Carla or AirSim~\cite{shah2018airsim}, the proposed solution is based on a customizable simulation framework, which offers the following advantages:
(i) high flexibility, since the application is not restricted to self-driving cars or drones, but can be implemented for any mobile robot;
(ii) custom  virtual environments, which can be designed by the user or easily downloaded from the Unreal marketplace;
(iii) efficiency, as \NAME~is fairly lightweight and does not require complex installations or painful migrations with future minor releases of Unreal Engine.

%% file: 03-method.tex
\section{Proposed Framework}\label{s:method}
This section describes the architecture of \NAME; then, it details the requirements on the physical robot side; finally, it illustrates the functionalities of the proposed framework.

\subsection{Framework architecture}
The proposed framework is based on ROS 2~\cite{ROS2Humble}, which provides an effective communication support between distinct computational nodes, even in distributed settings.  

A high-level overview of the \NAME~architecture is illustrated in Figure~\ref{fig:fig1}. \NAME\ subscribes to the topics published by the physical robot (requiring the two components to be connected to the same network). The physical robot must be equipped with a standard ROS 2 communication setup, while \NAME\ relies on the ROSIntegrationTool plugin~\cite{mania19scenarios} for communication, which allows publishing and subscribing to standard ROS 2 topics through dedicated callbacks. The plugin requires the ROS Bridge suite to support TCP packet exchange.

\subsection{Physical robot side}

Apart from being built on ROS 2, the only requirement on the physical robot side is related to the localization functionality: to accurately replicate its motion in the digital world, the robot must share its pose (position and orientation) in a dedicated topic. 
It is worth noting that the simulation framework returns synthetic sensory data to the robot in dedicated topics, different from the ones where the physical sensor data are published. The synthetic data are then used as input to the algorithm under test.
Depending on where the physical robot is placed and moved in the real world, collisions might occur. For this reason, the robot should have a safety stop mechanism to avoid collisions in the real world (however, not mandatory for the framework to operate).
When a safety stop occurs, it must be notified to \NAME, which will pause the simulation and resume it as soon as the robot is brought back to a safe position. The implementation of this mechanism is detailed in Section~\ref{ss:implementation}.

Another optional (but useful) boolean signal that the robot might want to communicate is the reset flag, which is useful whenever the simulation should be restarted (e.g., at the start of the simulation, when a virtual collision occurs, or when the task is completed).
Such a signal will trigger the simulation framework to re-initialize the environment.

\subsection{\NAME}\label{ss:ue5_side}


Figure \ref{fig:fig1} shows the main functional modules composing \NAME. The ROS Orchestrator module is responsible for receiving the messages from the topics through dedicated callbacks, which trigger different operational modes of the simulator. Such modes are encoded in specific functions of the other modules, i.e., the non-playing character (NPC) Spawner, the Object Spawner, and the Vehicle Manager.
The following paragraphs provide details of each callback.

\paragraph{\texttt{Pose} callback}
The \texttt{Pose} callback triggers the Vehicle Manager to update the digital twin's position in the virtual world, check collisions, and generate new sensory data.

\begin{figure}
    \centering
    \includegraphics[width=\linewidth]{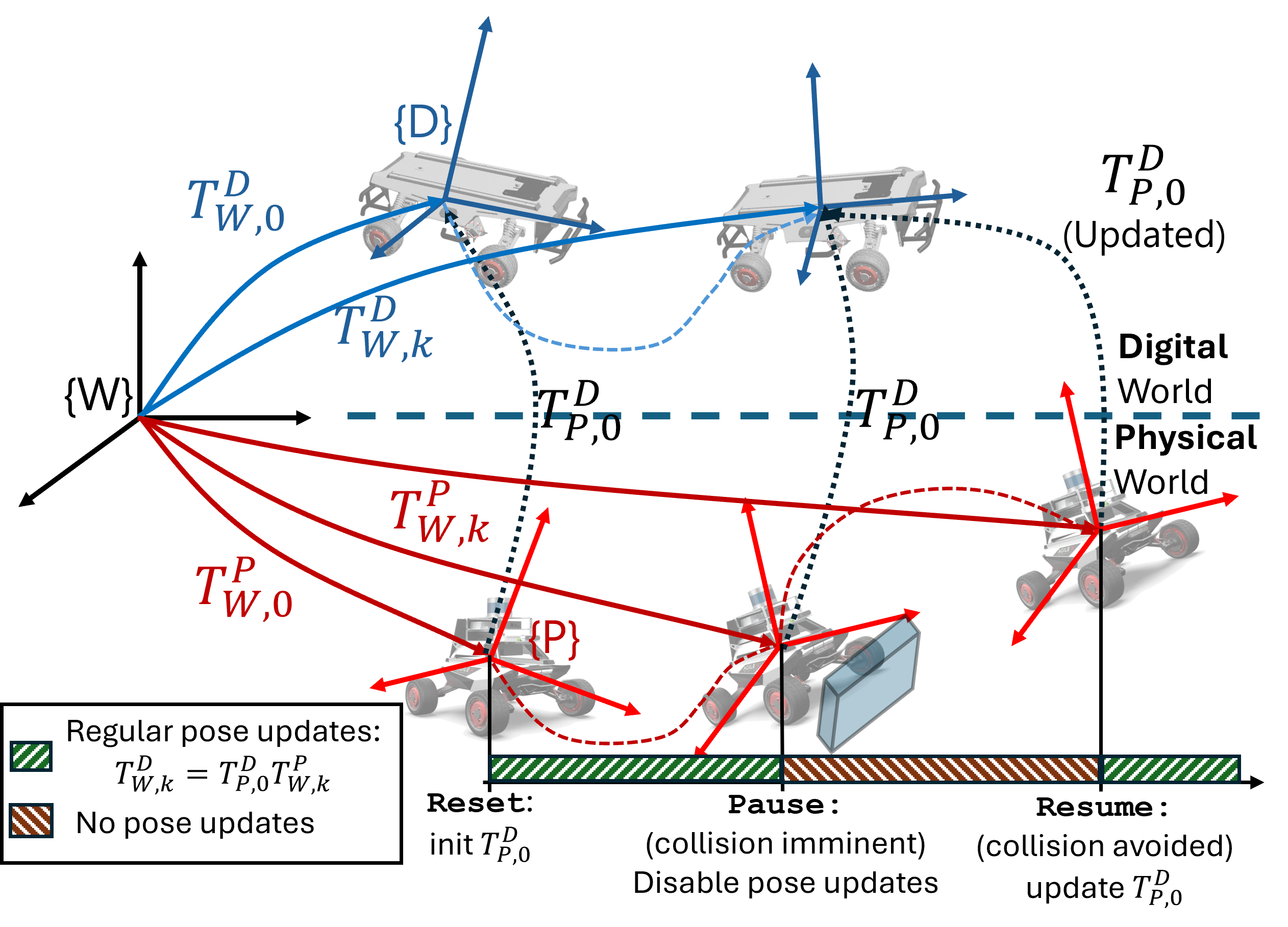}
    \caption{\small Definition of the Digital, Physical and World frames and corresponding transformations. The pose of the digital frame ($T_{W,k}^D$) is computed from the physical one ($T_{W,k}^P$) by transforming it with the offset $T_{P,0}^D$. The timeline on the bottom of the image illustrates the working principles of the \texttt{Pause}/\texttt{Resume} callbacks and how the offset is updated.} 
    \label{fig:frames}
\end{figure}

To correctly move the digital twin in the scene, the raw position message from the physical robot must be transformed to obtain meaningful poses in the digital domain. This requires the definition of three different frames, illustrated in Figure~\ref{fig:frames}: the body-frame $P$ placed on the physical robot, the body-frame $D$ placed on the digital twin, and an intermediate fixed world frame $W$ shared between the two domains.

The physical rover's pose can be expressed as a 4$\times$4 roto-translation matrix $T_W^P$ made up of the rotation matrix $R_W^P$ and the translation vector $\overline{WP}^P$. The same can be defined for the digital frame as $T_W^D$, made up of $R_W^D$ and $\overline{WD}^D$. Each of these quantities can be indexed with an additional subscript to indicate the corresponding timestep $k$, where $k=0$ denotes the timestep at initialization.

The initial physical robot's pose $\overline{WP}_0^W$ might not be feasible in the virtual world, and, in general, it will differ from the initial position of the digital twin. Hence, when initializing the position and rotation of the digital twin (when the \texttt{Reset} callback is triggered), it is necessary to compute a roto-translational offset $T_{P, 0}^D$ made up of $R_{P,0}^D$ and the corresponding translation
vector $\overline{PD}^D_0$,
which is then used at runtime to compute the correct digital position and rotation at timestep $k$ from the physical one as 

\begin{equation}
\begin{cases}
\overline{WD}_k^D = R_{P,0}^D \left(\overline{WP}^P_k - \overline{WP}_0^P\right) + \overline{WD}^D_0 \\
R_{W, k}^D=R_{P,0}^D R_{W,k}^P.
\end{cases}
\end{equation}\label{e:positioning}

The offset can be computed as $T_{P,0}^D=T_{W,0}^D \left(T_{W,0}^P\right)^{-1}$ and must be recomputed when the simulation is resumed after pausing. In fact, when the \texttt{Pause} callback is triggered, the \texttt{Pose} callback is disabled to allow the physical rover to move while its digital twin stays still. When resuming the simulation by triggering the dedicated callback, the \texttt{Pose} callback is re-enabled and the offset is updated to restart from the exact same position, while the physical robot moved in a different spot. This mechanism is illustrated in Figure \ref{fig:frames}. The complete equations are not reported for space limitations. 

After moving the digital twin, the Vehicle Manager checks collisions between the digital twin and other meshes in the virtual environment by controlling whether the collision volume of the robot's mesh overlaps with the collision volume of another entity.
This is a standard mechanism in virtual reality and it is convenient since the physics of the robot and its interactions with other virtual objects are disabled, as they could not be reproduced in the real world.
If there is no collision, the Vehicle Manager waits for the generation of the sensory data.
Sensors are coded as additional objects (not showed in Figure \ref{fig:fig1} for simplicity) that can be configured and attached to the robot.
When sensory data generation is complete, the ROS Orchestrator publishes them into the dedicated topics.

\paragraph{\texttt{Reset} Callback}
The \texttt{Reset} callback is triggered whenever the dedicated reset topic is published. 
The Vehicle Manager is responsible for the initialization of the position of the robot, which can be spawned in pre-defined areas of the virtual environment. 
The Object Spawner is called to manage the spawn of new (and destruction of old) static objects within pre-defined areas. 
Similarly, the Non-Playing Character (NPC) Spawner can be configured to spawn new dynamic objects (and destroy old ones) such as pedestrians or other robots. The area where the NPCs can be spawned and are allowed to move can be configured as well.

\paragraph{\texttt{Pause}/\texttt{Resume} Callback}
The \texttt{Pause} callback is triggered whenever the simulation must be paused. As mentioned above and illustrated in Figure \ref{fig:frames}, this may happen when a collision in the real world must be avoided. When the \texttt{Pause} callback is called, the real robot may move freely in the physical world without moving its digital twin. This is obtained by pausing the simulation and disabling the \texttt{Pose} callback.
When the robot is again in a safe position that allows resuming the simulation, it is possible to publish in the \texttt{Resume} callback, which updates the offset $T_{P, 0}^D$, re-enables the \texttt{Pose} callback, and resumes the simulation.

\subsection{Additional modes}

\NAME~can operate under different modes, according to the user's needs. Although it was initially developed for direct use with hardware (i.e., the physical robot), there are other cases where the simulation framework might result useful. They are described below.

\paragraph{Using a simulated robot} It might be a cheap and less cumbersome alternative to the use of a real one. Being built on ROS 2, the framework can seamlessly work also with simulated robots. For instance, Gazebo allows robot simulation with realistic models, but does not have a photo-realistic graphics engine: our simulation framework can provide photo-realism and virtual environments that can easily be edited. Hence, preliminary testing or fine-tuning of algorithms can be performed using Gazebo robot simulations augmented with the proposed UE5-based renderings.

\paragraph{Simulating the robot dynamics} If a real robot cannot be used and a Gazebo simulation is not available, the proposed simulation framework can be configured to simulate simple kinematic models that take velocity commands as inputs. This simulation mode is particularly useful when an AI agent (e.g., based on reinforcement learning) must be trained directly with photo-realistic renderings. The internal model reduces kinematics/dynamics realism to improve training efficiency, which is one of the main hurdles when training directly on a physical robot.

%% file: 04-experiments.tex
\section{Experimental Results}\label{s:exp}

This section describes some details about the implementation used to validate the simulation framework and presents the achieved results.

\subsection{Implementation details}\label{ss:implementation}

The framework was developed and tested with a distributed setup, where Unreal Engine 5.2 ran on a Windows 11 PC with an i9-9900 core, 32 GB of RAM, and an NVidia GeForce RTX 3070 GPU. 
Although the framework supports both real and simulated robots (on a different Ubuntu PC), all the experiments described here were performed with a physical AgileX Scout Mini rover.
The simulation framework in UE5 was developed in C++, hence all the UE5 objects described in Section~\ref{ss:ue5_side} (namely the ROS Orchestrator, the Vehicle Manager, and the Spawners) are C++ classes of the Actor type. Their properties are visible and editable directly from the UE5 Editor. A brief guide with configuration instructions will be released with the code.

Please note that, since UE5 uses centimeters as a base unit and left-hand rotations, all the dimensions have been scaled by a factor of 100 and the yaw and pitch values have been reversed.
Also, while the virtual world executes synchronously, the callback mechanism operates asynchronously. Therefore, the robot's position update rate strictly depends on the position topic publishing rate and the UE5 execution time. 

The following paragraphs provide details about the task simulated by \NAME
, the control software stack of the physical robot
, and 3D asset sources and licensing.

\paragraph{The task}\label{ss:task}
The proposed framework was validated on a task consisting of camera- and LiDAR-based corridor navigation and obstacle avoidance, where obstacles included various objects and stationary pedestrians.
More specifically, the LiDAR was used to navigate the corridor to avoid collisions, whereas the camera was used to detect pedestrians and reduce speed if required. A QR code was placed at the end of the corridor to indicate the end of the course. 
For demonstration purposes, the physical rover was placed in a small empty room (5m $\times$ 3m).

\begin{figure}
    \centering
    \includegraphics[width=0.7\linewidth]{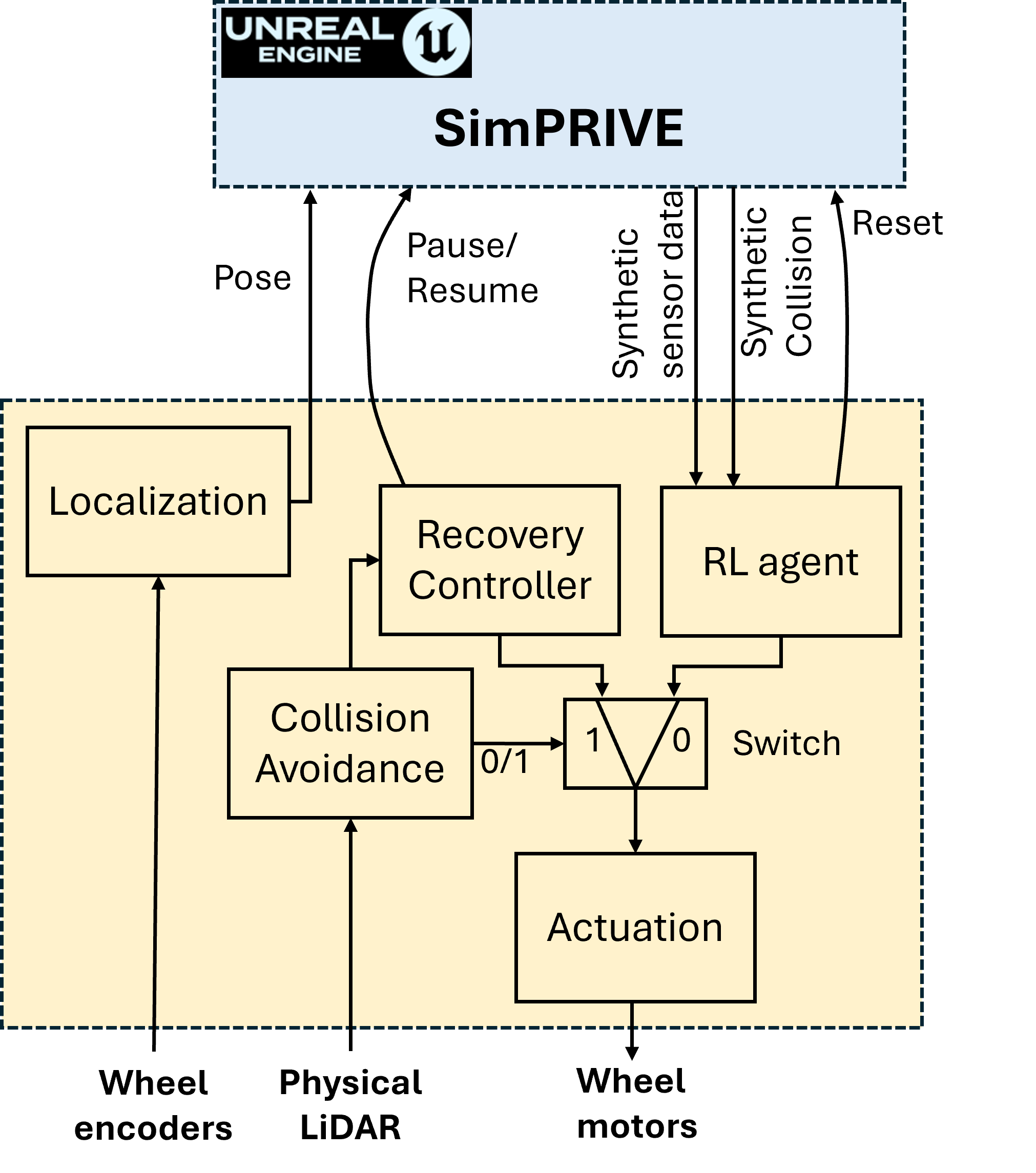}
    \caption{\small Functional control architecture of the rover on-board software used for the experiments. The arrows specify the dedicated \NAME\ ROS 2 topics.}
    \label{fig:impl_arch}
\end{figure}

\paragraph{Physical Robot}\label{ss:impl_robot}
The mobile robot is an AgileX Scout Mini rover~\cite{scout_mini}, equipped with an RGB camera with a 90-degree field of view and 640×480 resolution, a 3D LiDAR with a 360-degree horizontal and 30-degree vertical field of view (both with 1-degree angular resolution), and a Kria KR260 board running Ubuntu 22 and ROS 2 Humble. Figure~\ref{fig:impl_arch} illustrates the software architecture of the rover and the ROS 2 topics used to communicate with \NAME.
The rover is equipped with a localization module that estimates its pose with wheel encoders. 
Positioning accuracy is affected by errors that could be reduced by exploiting inertial data from an IMU sensor. However, since accurate positioning is not the focus of this paper, it is left as future work. 

The rover is programmed to publish a Reset flag into the framework's dedicated topic, which initializes the virtual rover position and the virtual world by spawning obstacles and pedestrians in the corridor. 
Such meshes can easily be added from the UE5 Editor to an asset library in the Object and NPC Spawner classes, respectively.

The rover is asked to solve the virtual corridor navigation task with a pre-trained RL agent.
The synthetic LiDAR point cloud is downsampled to obtain 3 range values, one frontal and the other two at a +/-30-degree angle with the first one. The three range values are used as input to a Deep Deterministic Policy Gradients (DDPG)~\cite{lillicrap2015continuous} actor (trained off-line on a simplified simulation platform), which returns the velocity and steering commands to the rover, which is then actuated with a skid-steer control. 
Such velocity commands are saturated to 1m/s of linear velocity and 0.5 rad/s of angular velocity. The linear velocity is reduced to 0.5m/s if a pedestrian is detected in the synthetic camera image. The person detector is a YOLO-v8 \cite{ultralytics_yolov8_2023} pre-trained on COCO.

The collision avoidance algorithm in the real world takes physical LiDAR data as input. By checking the minimum distance from the closest obstacle it is possible to stop the rover before it touches it. To guarantee that collisions are avoided it was necessary to calibrate the minimum distance threshold to make sure that, at the maximum velocity allowed (1m/s), the rover would break in time. This threshold resulted to be 1.5m to reliably avoid collisions.

When the safety stop occurs, the rover publishes into the dedicated \texttt{Pause} topic to suspend the simulation; while the simulation is paused, the rover can disable the RL agent's output and activate the Recovery Controller that turns the rover (angular velocity 0.5rad/s) until enough free space is detected ($>$ 2.5m in this specific case). Please note that in our setup, the small room where the rover was placed was empty, and such simple controllers were enough to avoid collisions. More complex scenarios might require more sophisticated solutions.
After rotating the rover, the Recovery Controller commands the last velocity that was published before stopping, then publishes into the \texttt{Resume} topic to continue the simulation and reactivates the RL output.

\paragraph{Sources and Licenses}\label{ss:licensing}
The experiments and figures presented in this paper have been created with the following content: \cite{fab_listing1, fab_listing2, fab_listing3, fab_listing4}.
The code of the framework will be released upon publication. Additional licensing details will be provided with the release.

\subsection{Results}

\paragraph{Testing in the virtual world}
The simulation framework was tested by running the architecture illustrated in Figure~\ref{fig:impl_arch} to solve the corridor navigation task, consisting in reaching the QR code with no collisions.

\begin{figure}
    \centering
    \begin{subfigure}[b]{\columnwidth}
    \includegraphics[trim=0 80 0 80, clip,width=\linewidth]{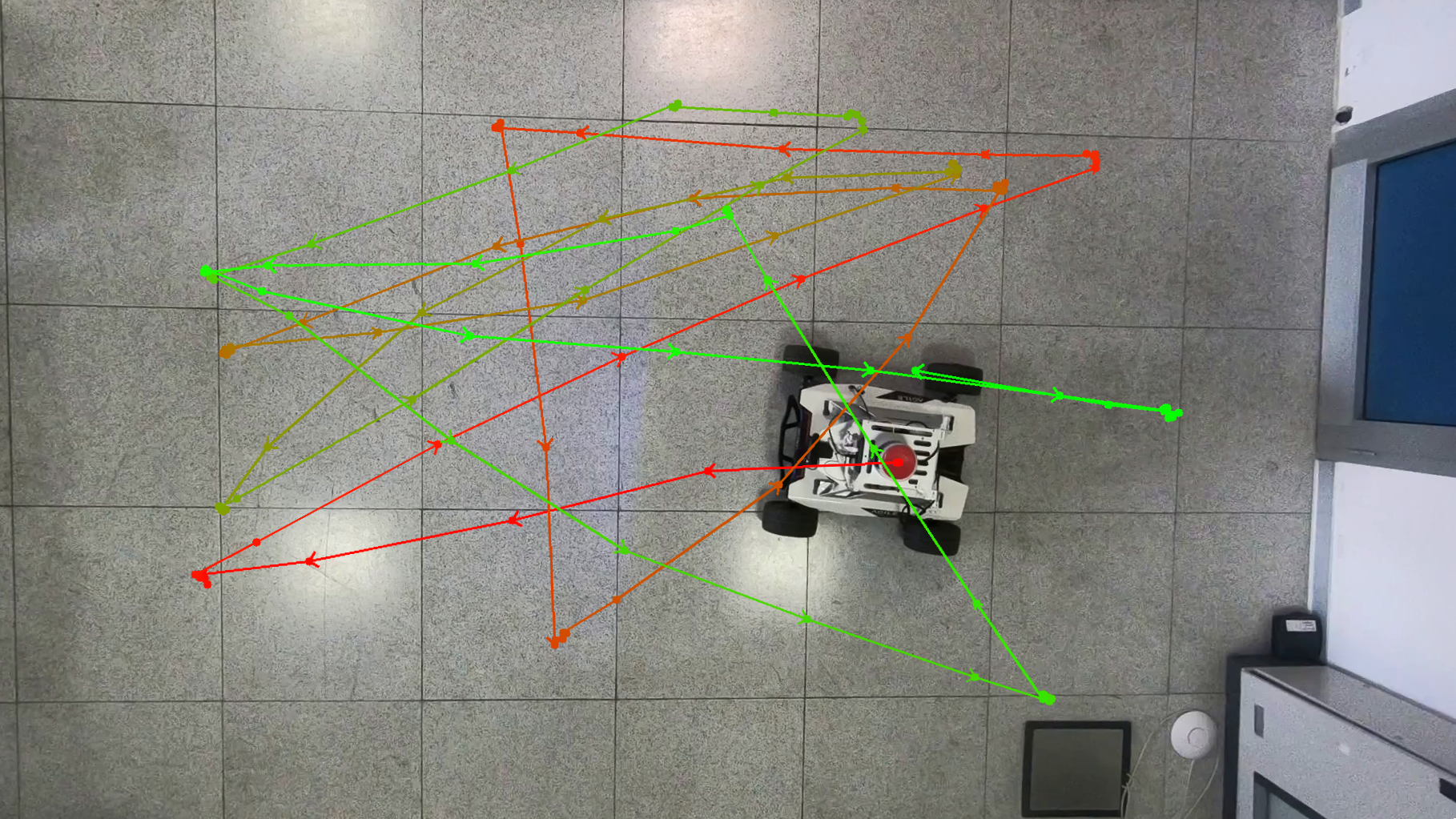}
    \end{subfigure}
    \begin{subfigure}[b]{\columnwidth}
    \includegraphics[trim=0 120 0 50, clip,width=\linewidth]{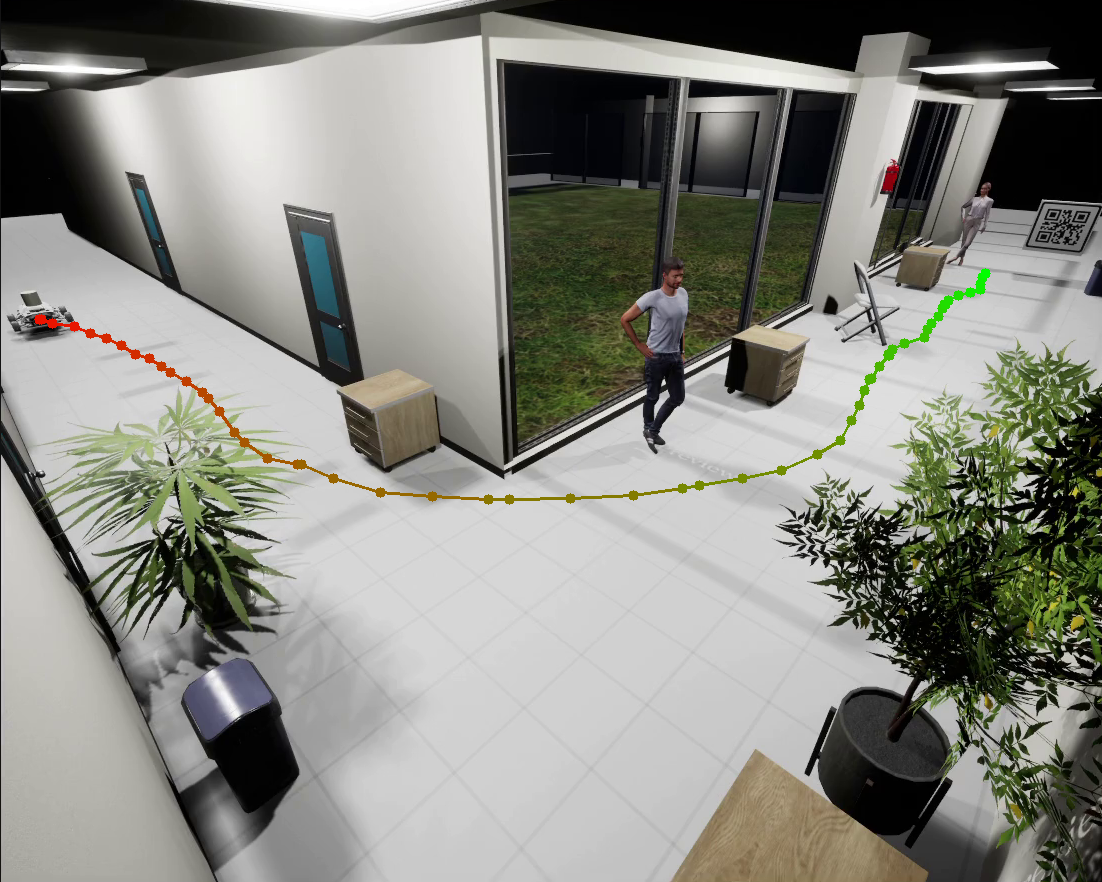}
    \end{subfigure}
    \caption{\small Path of the physical rover (top) and corresponding path of its digital twin (bottom). Each trajectory changes color from red (at the starting point) to green (at the final point) to better compare the positions in the physical and digital world.}
    \label{fig:exp1}
\end{figure}

Figure~\ref{fig:exp1} reports the trajectories of the physical rover and its digital twin. Each trajectory is drawn with a color that changes from red (at the starting point) to green (at the final point, to better compare the positions in the physical and digital world. Note that, while the trajectory of the digital twin is smooth and continuous, the one of the physical robot is interrupted abruptly whenever the rover gets too close to a wall. In such cases, the collision avoidance algorithm pauses the simulation, turns the rover (without changing the digital twin's position), and resumes the simulation. In this way, the entire rover's hardware and software stack (including the RL agent) is tested safely with no risk of collisions with physical obstacles.
The reduced space in the physical world allowed to stress-test the functionality of the framework, considering the large number of pause and resume required to cover the corridor in the virtual world.

\paragraph{Execution time}

\begin{figure}
    \centering
    \includegraphics[trim=20 0 40 35, clip,width=\linewidth]{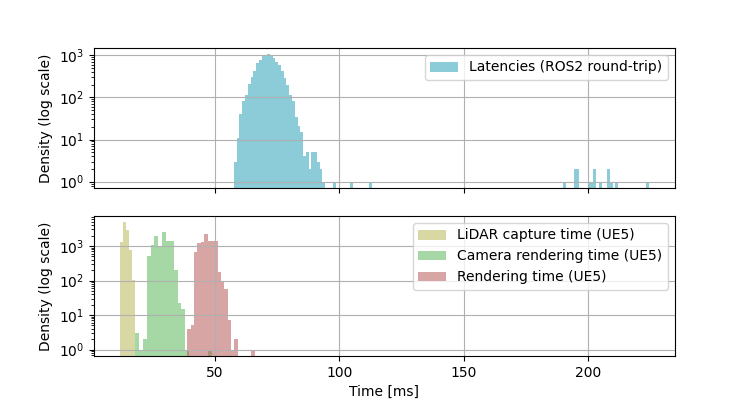}
    \caption{\small Distribution of the round-trip latency times (top) and the UE5 execution times for 10,000 iterations in log scale (bottom). While the rendering in UE5 shows predictable execution time, ROS2 occasionally introduces spurious delays. However, 99.82\% of the total measures result to be less than 100 ms.} 
    \label{fig:exp_timing}
\end{figure}

This experiment was carried out to evaluate (i) the time required by \NAME\ to move the digital twin and render synthetic sensor data, and (ii) the execution time of the entire simulation framework, including the communication latency, for a single loop.

On the physical rover side, a timer starts just before publishing the pose message that triggers the \texttt{Pose} callback and stops once all sensor and collision messages are received. In \NAME, the timer starts upon receiving a pose message and stops after publishing all virtual sensor and collision messages. These times were measured over 10,000 iterations. As expected, the sensor setup affects rendering time. Our setup is common in the field of robotics and it is sufficient for basic environmental perception. Different setups would affect framework performance, but UE5 supports asynchronous tasks and dedicated optimizations to reduce latency.
Figure~\ref{fig:exp_timing} shows the distributions of the round-trip time on the physical rover side (top) and the \NAME\ rendering and total execution time (bottom). The \NAME\ execution times are predictable, with an average of 47.04 ms, a standard deviation of 2.73 ms, and a worst-case of 66.00 ms (one occurrence). However, round-trip communication latency occasionally introduces delays over 200 ms, though this happened only 15 times out of 10,000 iterations. Minor delays resulted in round-trip times under 110 ms, with 99.82\% of measures below 100 ms, averaging 71.75 ms with a standard deviation of 6.72 ms. 
Overall, the combined rendering and communication latency is comparable to a typical 3D LiDAR acquisition (around 100 ms), making it suitable for real-time operations. Spurious large latencies can be mitigated by imposing a deadline on the physical rover actuation task, commanding a safety stop if a new message is delayed. These delays only affect the digital domain and are not critical.

%% file: 05-conclusion.tex
\section{Conclusions}\label{s:conclusions}
This paper presented \NAME, a flexible simulation framework for physical robot interactions with virtual environments, built in Unreal Engine 5 and ROS 2. The framework was designed to receive a pose from the physical robot and move its digital twin accordingly in a virtual environment. The position is never considered as absolute, but it is transformed to make it relative with respect to the initial position in the digital world. \NAME~also provides collision checks and sensor data generation in the virtual environment, which can be used by the physical robot as inputs for the algorithms under test.
The framework was validated by testing a reinforcement learning algorithm for obstacle avoidance, while the physical robot was placed in a confined space, using a LiDAR-based collision avoidance to safely navigate the physical world.

The current version of \NAME~has a few limitations that will be addressed in future updates. Firstly, configuring and customizing object behavior requires basic UE5 experience. This could be mitigated by providing a dedicated API, allowing users to access functionalities through a simpler language like Python, though this would limit customization flexibility.
Another issue is the simulation-to-reality gap due to differences between rendered and real images. Algorithms that work on synthetic images may not perform well on real images, especially if trained directly in simulation. Photo-realism depends on the quality of meshes in the virtual environment; high-quality meshes help bridge the gap between synthetic and real-world distributions. Mixed reality, which overlays virtual objects on real-world images, could address this issue but requires significant effort to render realistic lighting, shadows, occlusions, and accurate LiDAR data.
Finally, while \NAME~is designed for wheeled robots and can be extended to aerial vehicles, integrating other types of robots, such as quadrupeds or bipeds on uneven terrain, is more challenging. Legged robots rely heavily on ground contact forces, making flat surface simulation easier—a starting point for future extensions.

%% file: suppl.tex
\section{Supplementary Material}
\subsection{Computing the digital twin's pose}
As explained in Section XXX of the main paper, it is necessary to define the digital frame $D$ and the physical frame $P$. These two reference frames are fixed on the digital twin's mesh and on the physical robot, respectively. 

An intermediate World frame $W$ is required to refer the pose of both $D$ and $P$ to a fixed frame and to transform the quantities between the frames. 

The odometry message from the physical world reports the position $\overline{WP}^W_k$ and its rotation $R_{W,k}^P$. As explained in the main paper, the digital and physical domain are in general different, and the raw position of $P$ must be transformed according to a certain roto-translational offset, which is defined at the initialization of the simulation (when the Reset callback is triggered), i.e., at $k=0$.

The initial pose of the robot $T_{W,0}^D$ is known, as it comes from the spawn area allowed in the virtual world. In this notation, $T$ is a 4$\times$4 roto-translation matrix that encodes both rotation and translation. It is a typical formalism for robotics and mechanics:
\begin{equation*}
    T_A^B=
    \begin{bmatrix}
    R_A^B & \overline{AB}^B \\
    0_{1\times 3} & 1
    \end{bmatrix}
\end{equation*}

Therefore, at initialization it is possible to compute the offset
\begin{multline}
    T_{P,0}^D=T_{W,0}^D \left(T_{W,0}^P\right)^{-1} = \\
    \begin{bmatrix}
        R_{W,0}^D \left(R_{W,0}^P\right)^T & 
        \overline{WD}_0^D-R_{W,0}^D \left(R_{W,0}^P\right)^T \overline{WP}_0^P \\
        0_{1\times 3} & 1
    \end{bmatrix}
\end{multline}

that is used in the position update of the digital twin starting from the position $\overline{WP}_k^W$ and rotation $R_{W,k}^P$ of the physical rover (from the odometry message) :
\begin{multline}
    T_{W,k}^D=T_{P,0}^D T_{W,k}^P=\\
    \begin{bmatrix}
        R_{P,0}^D R_{W,k}^P & 
        \overline{WD}_0^D+R_{P,0}^D  R_{W,k}^P\left(\overline{WP}_k^W - \overline{WP}_0^W \right) \\
        0_{1\times 3} & 1
    \end{bmatrix}
\end{multline}

The offset must be recomputed whenever the simulation is paused and resumed. In this case, the offset is updated as 
\begin{equation}
    T_{P,0}^D=T_{W,k_{pause}}^D \left(T_{W,k_{resume}}^P\right)^{-1}
\end{equation}





\subsection{Additional illustrations}
\begin{figure}[h]
    \centering
    \includegraphics[width=\linewidth]{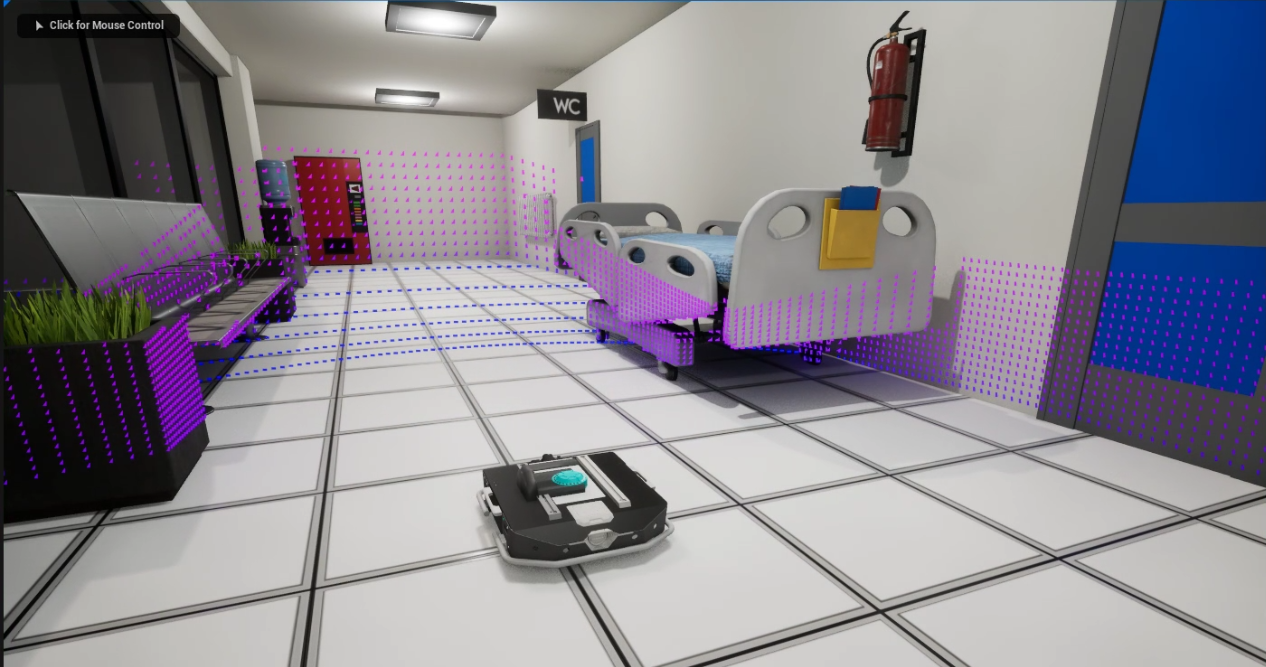}
    \caption{Illustration of the digital twin of the rover in the hospital environment.}
    \label{fig:suppl1}
\end{figure}

\begin{figure}[h]
    \centering
    \includegraphics[width=\linewidth]{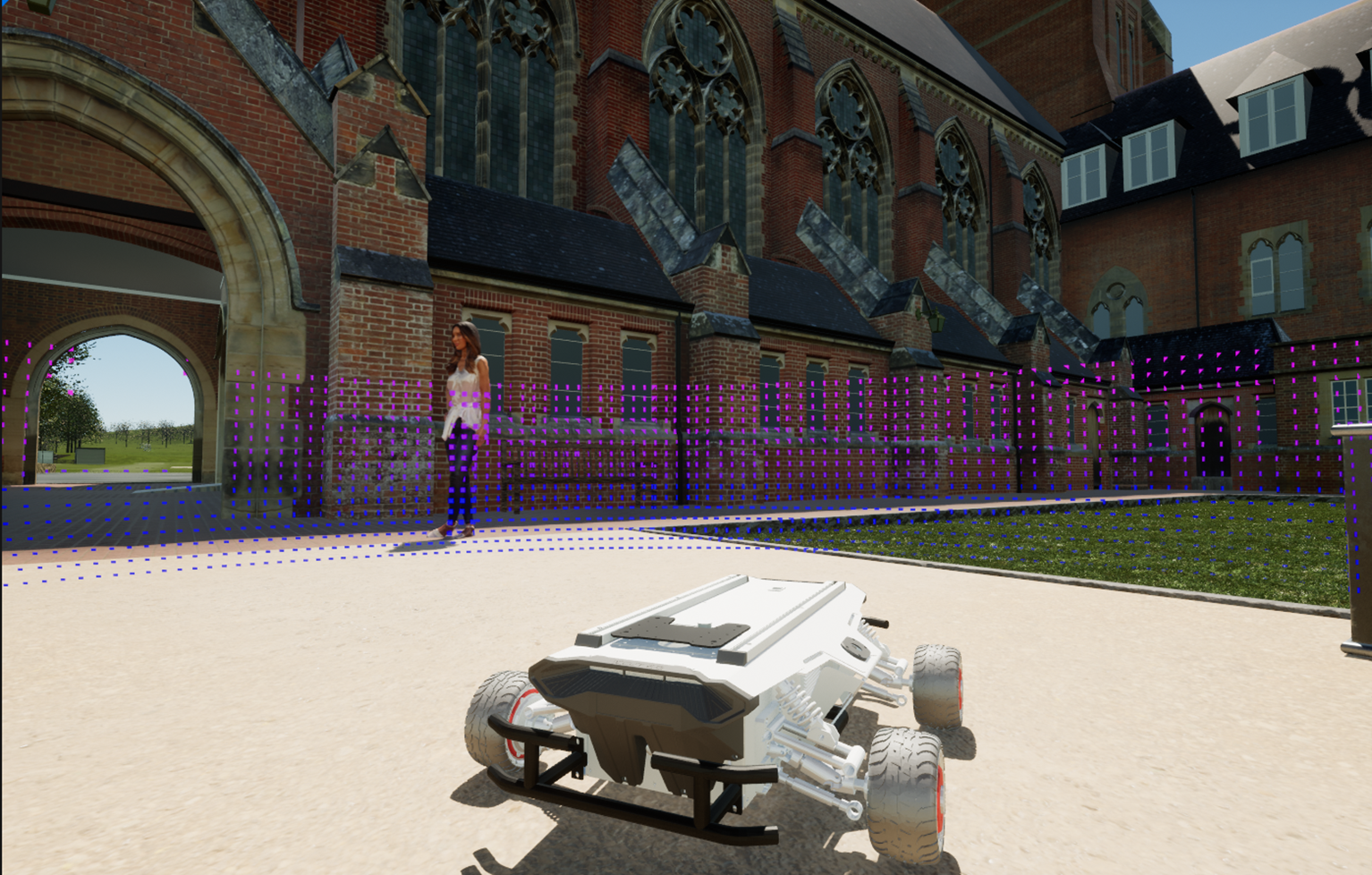}
    \caption{Illustration of the digital twin of the rover in the campus environment.}
    \label{fig:suppl2}
\end{figure}

\begin{figure}[h]
    \centering
    \includegraphics[width=\linewidth]{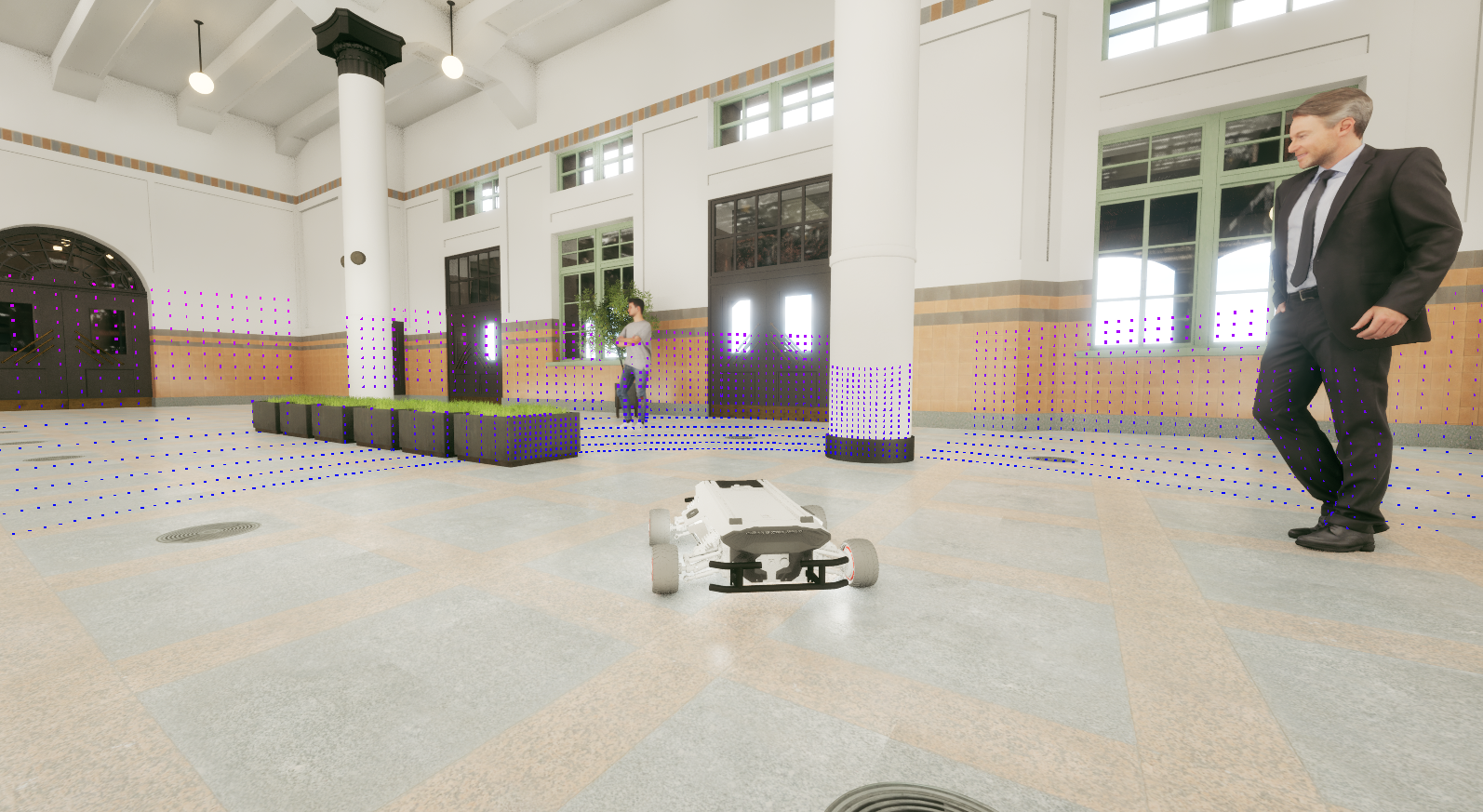}
    \caption{Illustration of the digital twin of the rover in the station environment.}
    \label{fig:suppl3}
\end{figure}

\begin{figure}[h]
    \centering
    \includegraphics[width=\linewidth]{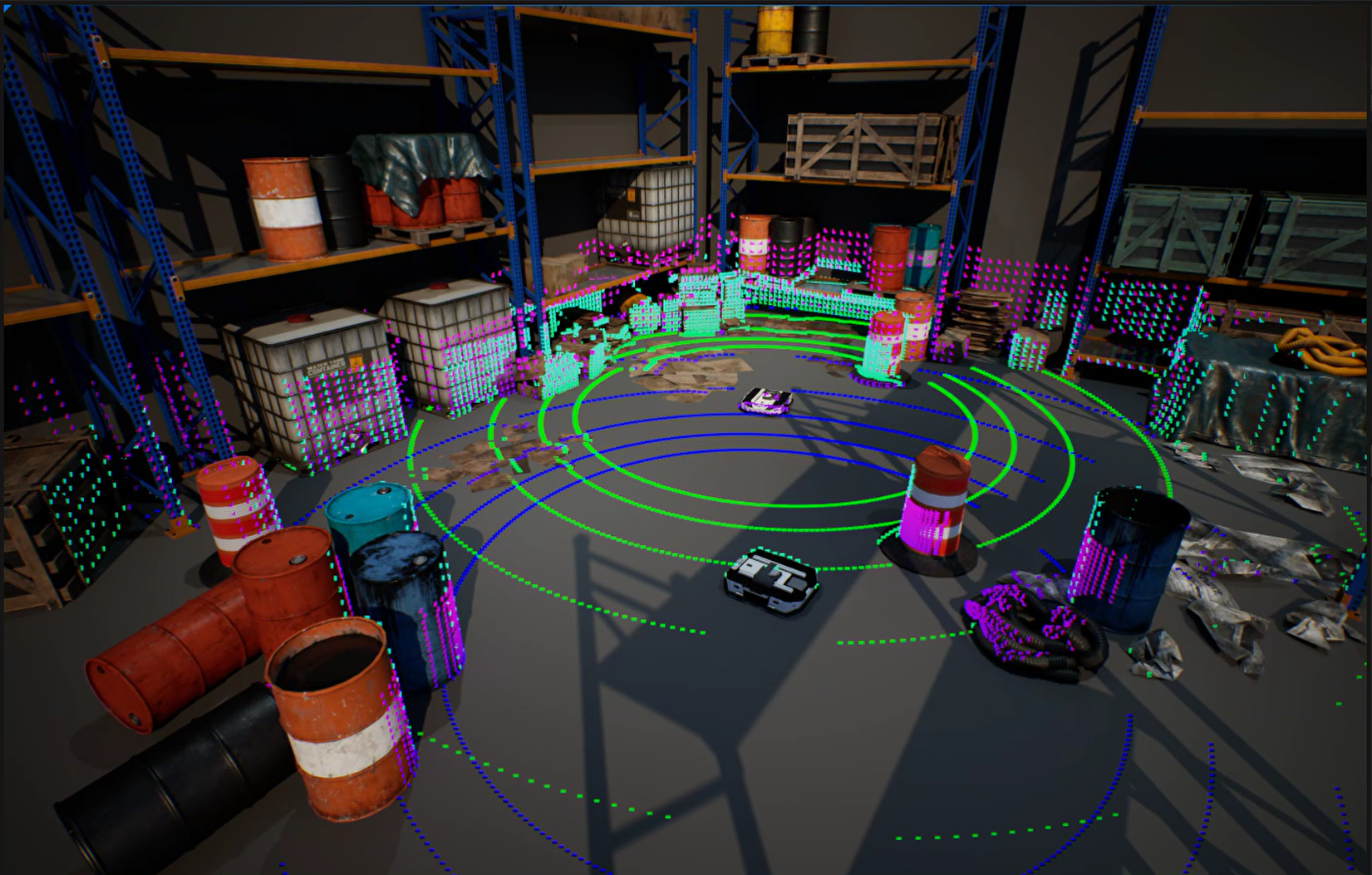}
    \caption{Illustration of the digital twin of the rover in the warehouse environment.}
    \label{fig:suppl4}
\end{figure}

This section provides some additional illustrations. \NAME~was tested in different virtual environments that might benefit from the application of mobile robotics: 
(i) a hospital/clinic (Figure~\ref{fig:suppl1};
(ii) a campus environment (Figure~\ref{fig:suppl2};
(iii) a station environment (Figure~\ref{fig:suppl3}; and
(iv) a warehouse environment (Figure~\ref{fig:suppl4}.

\subsection{Training the Reinforcement Learning agent}
The DDPG agent was trained on a simplified simulation environment. Since the agent relies on LiDAR only to navigate the corridor (while the camera is used mainly to detect people and the finish QR code), a restricted geometric model is enough to make the agent learn the correct behavior.

The environment is initialized by computing randomized trajectories, created by drawing random commands for the dynamical model of a unicycle. Such trajectory is then used as the center line to calculate the positions of the corridor walls with fixed width. Randomized obstacles (squares) are placed on the sides of the corridor to obtain a new maze for each episode.
The rover is simulated with a unicycle kinematics model and its footprint is approximated with a square with side 1m. A collision occurs as soon as one of the sides of the square intersects any of the sides of the obstacles or any of the borders.

The LiDAR is simulated with 2D ray-tracing and checking the intersections with the obstacles and the borders. Its maximum range is assumed to be 10m.

As stated in the main paper, the agent takes as input 3 rays ${d_f, d_r, d_l}$ from the LiDAR ($d_f$ the frontal one and $d_r$ and $d_l$ are the two rays at $\pm$30 degrees, right and left respectively) and outputs the linear and the angular velocity that are used to control the rover. 

The reward for the agent is defined as: -1 for each step; +0.1$d_f$ to promote the frontal ray to be obstacle free; -0.1$|d_r -d_l|$ to balance the free space between the left and right rays; the agent also obtains +10 every 5 meters traveled forward toward the goal, +100 when reaching the goal, and -100 for each collision. 

Both the critic and the actor networks are multi-layer perceptrons with 2 hidden layers (300 neurons in the first hidden layer and 400 neurons in the second) with ReLU activations. The only non-ReLU activation is the tanh on the output, which effectively restricts the output between -1 and 1. Then, the angular velocity is scaled by a factor of 0.5.

The agent is trained using the code in the repository \url{https://github.com/ghliu/pytorch-ddpg}, keeping the default hyperparameters.

%% file: main.bbl
\begin{thebibliography}{10}
\providecommand{\url}[1]{#1}
\csname url@samestyle\endcsname
\providecommand{\newblock}{\relax}
\providecommand{\bibinfo}[2]{#2}
\providecommand{\BIBentrySTDinterwordspacing}{\spaceskip=0pt\relax}
\providecommand{\BIBentryALTinterwordstretchfactor}{4}
\providecommand{\BIBentryALTinterwordspacing}{\spaceskip=\fontdimen2\font plus
\BIBentryALTinterwordstretchfactor\fontdimen3\font minus \fontdimen4\font\relax}
\providecommand{\BIBforeignlanguage}[2]{{%
\expandafter\ifx\csname l@#1\endcsname\relax
\typeout{** WARNING: IEEEtran.bst: No hyphenation pattern has been}%
\typeout{** loaded for the language `#1'. Using the pattern for}%
\typeout{** the default language instead.}%
\else
\language=\csname l@#1\endcsname
\fi
#2}}
\providecommand{\BIBdecl}{\relax}
\BIBdecl

\bibitem{rech2024artificial}
P.~Rech, ``Artificial neural networks for space and safety-critical applications: Reliability issues and potential solutions,'' \emph{IEEE Transactions on Nuclear Science}, vol.~71, no.~4, pp. 377--404, 2024.

\bibitem{perez2024artificial}
J.~Perez-Cerrolaza, J.~Abella, M.~Borg, C.~Donzella, J.~Cerquides, F.~J. Cazorla, C.~Englund, M.~Tauber, G.~Nikolakopoulos, and J.~L. Flores, ``Artificial intelligence for safety-critical systems in industrial and transportation domains: A survey,'' \emph{ACM Computing Surveys}, vol.~56, no.~7, pp. 1--40, 2024.

\bibitem{8897630}
A.~Biondi, F.~Nesti, G.~Cicero, D.~Casini, and G.~Buttazzo, ``A safe, secure, and predictable software architecture for deep learning in safety-critical systems,'' \emph{IEEE Embedded Systems Letters}, vol.~12, no.~3, pp. 78--82, 2020.

\bibitem{wang2024survey}
J.~Wang, J.~Ai, M.~Lu, H.~Su, D.~Yu, Y.~Zhang, J.~Zhu, and J.~Liu, ``A survey of neural network robustness assessment in image recognition,'' \emph{arXiv preprint arXiv:2404.08285}, 2024.

\bibitem{meng2022adversarial}
M.~H. Meng, G.~Bai, S.~G. Teo, Z.~Hou, Y.~Xiao, Y.~Lin, and J.~S. Dong, ``Adversarial robustness of deep neural networks: A survey from a formal verification perspective,'' \emph{IEEE Transactions on Dependable and Secure Computing}, 2022.

\bibitem{aldahdooh2022adversarial}
A.~Aldahdooh, W.~Hamidouche, S.~A. Fezza, and O.~D{\'e}forges, ``Adversarial example detection for dnn models: A review and experimental comparison,'' \emph{Artificial Intelligence Review}, vol.~55, no.~6, pp. 4403--4462, 2022.

\bibitem{yang2024generalized}
J.~Yang, K.~Zhou, Y.~Li, and Z.~Liu, ``Generalized out-of-distribution detection: A survey,'' \emph{International Journal of Computer Vision}, vol. 132, no.~12, pp. 5635--5662, 2024.

\bibitem{salehi2021unified}
M.~Salehi, H.~Mirzaei, D.~Hendrycks, Y.~Li, M.~H. Rohban, and M.~Sabokrou, ``A unified survey on anomaly, novelty, open-set, and out-of-distribution detection: Solutions and future challenges,'' \emph{arXiv preprint arXiv:2110.14051}, 2021.

\bibitem{unity}
\BIBentryALTinterwordspacing
{Unity Technologies}, ``Unity,'' 2023, game development platform. [Online]. Available: \url{https://unity.com/}
\BIBentrySTDinterwordspacing

\bibitem{ue5}
\BIBentryALTinterwordspacing
{Epic Games}, ``Unreal engine 5,'' 2025, game development platform. [Online]. Available: \url{https://www.unrealengine.com/en-US/unreal-engine-5}
\BIBentrySTDinterwordspacing

\bibitem{10576049}
F.~Nesti, G.~Rossolini, G.~D’Amico, A.~Biondi, and G.~Buttazzo, ``Carla-gear: A dataset generator for a systematic evaluation of adversarial robustness of deep learning vision models,'' \emph{IEEE Transactions on Intelligent Transportation Systems}, vol.~25, no.~8, pp. 9840--9851, 2024.

\bibitem{mihalivc2022hardware}
F.~Mihali{\v{c}}, M.~Trunti{\v{c}}, and A.~Hren, ``Hardware-in-the-loop simulations: A historical overview of engineering challenges,'' \emph{Electronics}, vol.~11, no.~15, p. 2462, 2022.

\bibitem{tao2022digital}
F.~Tao, B.~Xiao, Q.~Qi, J.~Cheng, and P.~Ji, ``Digital twin modeling,'' \emph{Journal of Manufacturing Systems}, vol.~64, pp. 372--389, 2022.

\bibitem{cheng2023survey}
\BIBentryALTinterwordspacing
J.~Cheng, Z.~Wang, X.~Zhao, Z.~Xu, M.~Ding, and K.~Takeda, ``A survey on testbench-based vehicle-in-the-loop simulation testing for autonomous vehicles: Architecture, principle, and equipment,'' \emph{Advanced Intelligent Systems}, vol. n/a, no. n/a, p. e202300778, 2023. [Online]. Available: \url{https://advanced.onlinelibrary.wiley.com/doi/pdfdirect/10.1002/aisy.202300778}
\BIBentrySTDinterwordspacing

\bibitem{scout_mini}
\BIBentryALTinterwordspacing
{AgileX}, ``Agilex scout mini,'' 2025. [Online]. Available: \url{https://global.agilex.ai/products/scout-mini}
\BIBentrySTDinterwordspacing

\bibitem{nasa_digital_twin}
B.~D. Allen, ``Digital twins and living models at nasa,'' \url{https://ntrs.nasa.gov/citations/20210023699}, 2021, accessed: 2025-03-31.

\bibitem{9670454}
C.~Schwarz and Z.~Wang, ``The role of digital twins in connected and automated vehicles,'' \emph{IEEE Intelligent Transportation Systems Magazine}, vol.~14, no.~6, pp. 41--51, 2022.

\bibitem{botin2022digital}
\BIBentryALTinterwordspacing
D.~M. Botín-Sanabria, A.-S. Mihaita, R.~E. Peimbert-García, M.~A. Ramírez-Moreno, R.~A. Ramírez-Mendoza, and J.~de~J.~Lozoya-Santos, ``Digital twin technology challenges and applications: A comprehensive review,'' \emph{Remote Sensing}, vol.~14, no.~6, p. 1335, 2022. [Online]. Available: \url{https://www.mdpi.com/2072-4292/14/6/1335}
\BIBentrySTDinterwordspacing

\bibitem{9361649}
Z.~Szalay, ``Next generation x-in-the-loop validation methodology for automated vehicle systems,'' \emph{IEEE Access}, vol.~9, pp. 35\,616--35\,632, 2021.

\bibitem{shen2023simonwheels}
Y.~Shen, B.~Chandaka, Z.-H. Lin, A.~Zhai, H.~Cui, D.~Forsyth, and S.~Wang, ``Sim-on-wheels: Physical world in the loop simulation for self-driving,'' vol.~8, no.~12, 2023, pp. 8192--8199.

\bibitem{10057099}
Z.~Wang, O.~Zheng, L.~Li, M.~Abdel-Aty, C.~Cruz-Neira, and Z.~Islam, ``Towards next generation of pedestrian and connected vehicle in-the-loop research: A digital twin co-simulation framework,'' \emph{IEEE Transactions on Intelligent Vehicles}, vol.~8, no.~4, pp. 2674--2683, 2023.

\bibitem{dosovitskiy2017carla}
A.~Dosovitskiy, G.~Ros, F.~Codevilla, A.~Lopez, and V.~Koltun, ``Carla: An open urban driving simulator,'' in \emph{Conference on robot learning}.\hskip 1em plus 0.5em minus 0.4em\relax PMLR, 2017, pp. 1--16.

\bibitem{xiong2022design}
H.~Xiong, Z.~Wang, G.~Wu, and Y.~Pan, ``Design and implementation of digital twin-assisted simulation method for autonomous vehicle in car-following scenario,'' \emph{Journal of Sensors}, vol. 2022, no.~1, p. 4879490, 2022.

\bibitem{wang2021digital}
Z.~Wang, K.~Han, and P.~Tiwari, ``Digital twin simulation of connected and automated vehicles with the unity game engine,'' in \emph{2021 IEEE 1st International Conference on Digital Twins and Parallel Intelligence (DTPI)}.\hskip 1em plus 0.5em minus 0.4em\relax IEEE, 2021, pp. 1--4.

\bibitem{VOOGD20231510}
\BIBentryALTinterwordspacing
K.~L. Voogd, J.~P. Allamaa, J.~Alonso-Mora, and T.~D. Son, ``Reinforcement learning from simulation to real world autonomous driving using digital twin,'' \emph{IFAC-PapersOnLine}, vol.~56, no.~2, pp. 1510--1515, 2023, 22nd IFAC World Congress. [Online]. Available: \url{https://www.sciencedirect.com/science/article/pii/S2405896323022553}
\BIBentrySTDinterwordspacing

\bibitem{allamaa2022sim2real}
J.~P. Allamaa, P.~Patrinos, H.~Van~der Auweraer, and T.~D. Son, ``Sim2real for autonomous vehicle control using executable digital twin,'' \emph{IFAC-PapersOnLine}, vol.~55, no.~24, pp. 385--391, 2022.

\bibitem{10156429}
A.~Hiba, V.~Kortvelyesi, A.~Kiskaroly, O.~Bhoite, P.~David, and A.~Majdik, ``Indoor vehicle-in-the-loop simulation of unmanned micro aerial vehicle with artificial companion,'' in \emph{2023 International Conference on Unmanned Aircraft Systems (ICUAS)}, 2023, pp. 137--143.

\bibitem{shah2018airsim}
S.~Shah, D.~Dey, C.~Lovett, and A.~Kapoor, ``Airsim: High-fidelity visual and physical simulation for autonomous vehicles,'' in \emph{Field and Service Robotics: Results of the 11th International Conference}.\hskip 1em plus 0.5em minus 0.4em\relax Springer, 2018, pp. 621--635.

\bibitem{ROS2Humble}
O.~Robotics, ``Ros 2 humble hawksbill,'' \url{https://docs.ros.org/en/humble/index.html}, 2022.

\bibitem{mania19scenarios}
P.~Mania and M.~Beetz, ``A framework for self-training perceptual agents in simulated photorealistic environments,'' in \emph{International Conference on Robotics and Automation (ICRA)}, Montreal, Canada, 2019.

\bibitem{lillicrap2015continuous}
T.~Lillicrap, ``Continuous control with deep reinforcement learning,'' \emph{arXiv preprint arXiv:1509.02971}, 2015.

\bibitem{ultralytics_yolov8_2023}
\BIBentryALTinterwordspacing
G.~Jocher, A.~Chaurasia, and J.~Qiu, ``Ultralytics yolov8,'' 2023. [Online]. Available: \url{https://github.com/ultralytics/ultralytics}
\BIBentrySTDinterwordspacing

\bibitem{fab_listing1}
M.~G. Art, ``Modular 3d hospital environment,'' \url{https://www.fab.com/listings/7e1574fd-1d28-4c6f-a612-c889c09078ef}, 2025, accessed: 2025-04-28.

\bibitem{fab_listing2}
J.~Assets, ``Grocery store props collection,'' \url{https://www.fab.com/listings/309d1733-ed82-4b57-b3fc-70ec26dc0641}, 2025, accessed: 2025-04-28.

\bibitem{fab_listing3}
AccuCities, ``English college level 4 sample,'' \url{https://www.fab.com/listings/374fb588-5711-41f1-a69b-4e60c95beea5}, 2025, accessed: 2025-04-28.

\bibitem{fab_listing4}
SilverTim, ``Industry props pack 6,'' \url{https://www.fab.com/listings/b5603e44-e1b0-4346-9c3d-04887aa9f87d}, 2025, accessed: 2025-04-28.

\end{thebibliography}
